\documentclass[acmsmall,screen,nonacm,review=false,timestamp=false]{acmart}
\AtBeginDocument{%
  }

\acmConference[IntRob'25]{Intelligent Robotics FAIR 2025}{June 23--24, 2025}{Budapest, Hungary}
\usepackage{svg}
\usepackage{float}    % Added for [H] placement specifier
\usepackage{enumitem}

\begin{document}

\title{Fast Real-Time Pipeline for Robust Arm Gesture Recognition}

\begin{center}
	Preprint
\end{center}

\author{Milán Zsolt Bagladi}
\email{hhpw8b@inf.elte.hu}
\affiliation{%
  \institution{ELTE Eötvös Loránd University Faculty of Informatics}
  \city{Budapest}
  \country{Hungary}
}

\author{László Gulyás}
\email{lgulyas@inf.elte.hu}
\affiliation{%
  \institution{ELTE Eötvös Loránd University Faculty of Informatics}
  \city{Budapest}
  \country{Hungary}
}

\author{Gergő Szalay}
\email{d5ij3p@inf.elte.hu}
\affiliation{%
  \institution{ELTE Eötvös Loránd University Faculty of Informatics}
  \city{Budapest}
  \country{Hungary}
}

%\renewcommand{\shortauthors}{Trovato et al.} - ezt kikommenteltem

\begin{abstract}
This paper presents a real-time pipeline for dynamic arm gesture recognition based on OpenPose keypoint estimation, keypoint normalization, and a recurrent neural network classifier. The \textit{$1 \times 1$} normalization scheme and two feature representations (coordinate- and angle-based) are presented for the pipeline. In addition, an efficient method to improve robustness against camera angle variations is also introduced by using artificially rotated training data.  Experiments on a custom traffic-control gesture dataset demonstrate high accuracy across varying viewing angles and speeds. Finally, an approach to calculate the speed of the arm signal (if necessary) is also presented.
% Finally, the approach is validated by an application where a TurtleBot 3 is controlled by arm signals.
\end{abstract}




%\received{15 May 2025}

\maketitle

%Main content
\section{Introduction}
\label{s:introduction}
Human arm gestures are frequently encountered in everyday life and are naturally interpretable by humans. However, processing these signals poses a significantly more complex challenge for computers. The recognition and interpretation of human gestures involve numerous difficulties, as the diversity of signals, the variability of environmental factors, and individual differences in movement all influence the accuracy of recognition.

In road traffic, it is crucial for autonomous vehicles to recognize police officers' hand signals used for directing traffic. Similarly, railway transport employs various hand signals to regulate train movements.  In aviation, ground crew hand signals guide aircraft safely at airports. The automatic recognition of such signals is needed to enhance the safety and efficiency of autonomous vehicles. In a broader context, the recognition and interpretation of natural gestures, including arm signals, may prove to be an important foundation for human-robot interactions, enabling  effective and intuitive communication between people and machines. 

This paper presents a novel approach to dynamic arm gesture recognition using a combination of keypoint estimation, a normalization technique, and recurrent neural networks. Our method achieves real-time performance with high accuracy. The method is robust against varying camera angles and varying gesture speeds. The  approach is evaluated on a custom dataset of traffic control gestures. %Its practical applicability is further demonstrated via a gesture-controlled mobile robot.

\section{Problem Statement}
\label{s:problem_statement}

Human arm gestures can be categorized into two main types: static and dynamic gestures. Static gestures can be recognized from a single frame, as one image contains all the necessary information for identification. In contrast, dynamic gestures cannot be identified from a single frame since their meaning is conveyed through movement, requiring temporal context for accurate interpretation. Importantly, dynamic gestures that pass through similar static poses can convey different meanings—for example, clockwise and counterclockwise circular motions share similar intermediate poses, but their temporal order distinguishes their respective meanings.

Besides the gesture path, its speed may also change the meaning of the signal. For instance, a quick circular motion may signal one command, while slow circling may mean another. In real settings, the person giving the signal might not face the camera directly, so the system must also recognize gestures seen at an angle.

The requirement for real-time recognition further complicates the problem, especially in applications such as autonomous vehicles or automated robots. In these domains, delayed or inaccurate recognition may lead to critical errors, making reliable, fast, and precise interpretation methods a necessity. % essential alongside correct signal identification.

This paper proposes an efficient and robust solution (pipeline) for automatic and real-time interpretation of dynamic arm gestures. This solution addresses the key challenges of dynamic gesture recognition. The differentiation of various gesture speeds is also considered.

\section{Related Work}
\label{s:related_work}
Recent studies leverage keypoint extraction and deep learning for gesture recognition in traffic and robotic applications. Bagladi \cite{tdk2024} uses OpenPose for keypoints, K-means and neural networks for static gestures, and a GRU for dynamic ones. Liu \emph{et al.} \cite{app112411951} employ an ST-GCN with adaptive graph structure and CBAM/TAS attention, reaching 87.7 \% top-1 on a 20 480-sample police dataset. Mishra \emph{et al.} \cite{s21237914} detect controllers with an object detector, reconstruct 3D hands via FrankMocap+SMPL-X, and classify with a CNN. Sathya \emph{et al.} \cite{SATHYA20151700} apply cumulative frame differences and block-motion features (CBIV) with Random Forest on a 25-subject static set. Ma \emph{et al.} \cite{ijgi7010037} build a real-time ST-CNN on a Kinect-based 155 000-frame TPCGS database, achieving 93.0 \% in virtual city tests. He \emph{et al.} \cite{HE2020248} combine a modified CPM, hand-crafted spatial features, and LSTM, obtaining 91.2 \% offline and 93.3 \% real-time accuracy, at the cost of separate training stages and higher compute demands. While these methods demonstrate strong performance, many do not address robustness to continuous variations in viewing angles or the potential significance of gesture execution speed, motivating the development of our proposed pipeline.

\section{The Proposed Pipeline}
\label{s:methodology}
\begin{figure}[htbp]
	\centering
	\includegraphics[width=1\columnwidth]{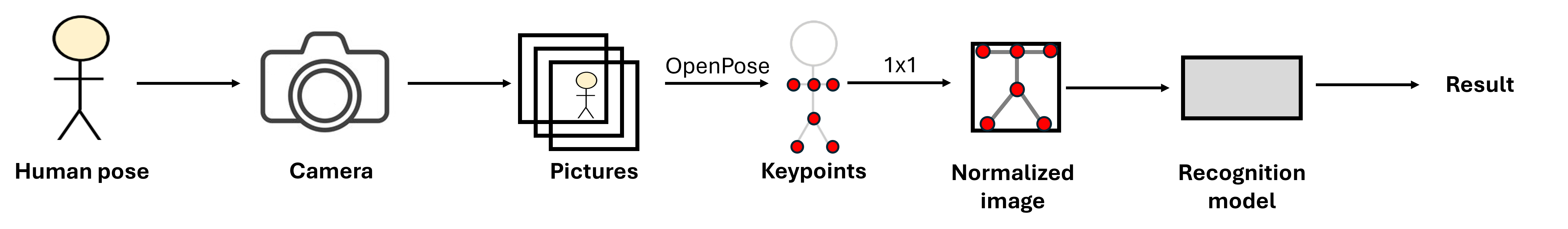}
	\caption{Pipeline overview}
 \label{fig:pipeline}
\end{figure}

This section introduces a pipeline for robust real-time interpretation of dynamic arm gestures (see Figure~\ref{fig:pipeline}). Our approach explicitly models the spatial direction, execution speed, and temporal progression of each gesture to ensure robust performance across diverse conditions and viewpoints. The system is structured into three main stages:
\begin{enumerate}
  \item \textbf{Keypoint Estimation:} Extract skeletal landmarks from each input frame.
  \item \textbf{Feature Extraction and Normalization:} Transform raw keypoints into invariant, informative descriptors.
  \item \textbf{Classification:} Employ a recurrent neural network to produce real-time gesture predictions.
\end{enumerate}

\subsection{Keypoint Estimation}
\label{ss:keypoint_estimation}
OpenPose is used for keypoint estimation. OpenPose is an open-source framework offering multiple keypoint detection models. \cite{cao2019openposerealtimemultiperson2d}
%We reconstruct a skeleton from detected keypoints. This approach meets our requirements and reduces the problem to numerical data processing, making it an appropriate choice.
Among these, the BODY-25 model is used as part of the pipeline. (BODY-25 is recommended for general applications as it provides a more detailed upper-body skeleton than the COCO model.) OpenPose outperforms more rudimentary pose estimators by being both fast and accurate. It includes pre-trained weights, effectively serving as a ready-to-use keypoint detector.

OpenPose is easily integrated into real-time pose estimation tasks thanks to its exceptional speed and precision. It is able to handle complex scenarios with minimal runtime overhead—a critical factor for live data processing. Additionally, OpenPose supports pre-trained models that require no further training and can detect multiple people simultaneously, broadening its applicability.

\subsection{Normalization}
\label{ss:normalization}

Raw OpenPose output is influenced by the position of the subject in the frame, motion toward or away from the camera, and variations in body shape. These factors result in inconsistent keypoint distributions, so the raw coordinates alone are insufficient for reliable gesture recognition.

Therefore, the raw OpenPose output is standardized. The goal is to ensure that poses perceived as similar by human observers remain close in the transformed keypoint space. Two poses are considered similar if an average person would judge them identical in all essential aspects; this perceived similarity is fundamental to our recognition process.

\begin{figure}[htbp]
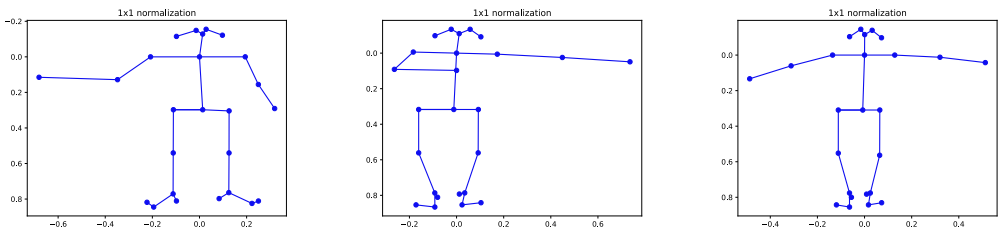

  \centering
  \includesvg[width=0.32\columnwidth]{images/1x1normapelda_1_ENG.svg}
  \hfill
  \includesvg[width=0.32\columnwidth]{images/1x1normapelda_2_ENG.svg}
  \hfill
  \includesvg[width=0.32\columnwidth]{images/1x1normapelda_3_ENG.svg} 
  \caption{Three examples of \textit{$1 \times 1$} normalization. Note that the essential information for gesture recognition remains intact after the transformation.}
  \label{fig:1x1normapelda}
\end{figure}

Positional dependence is removed by translating the keypoints so that a fixed reference point (BODY-25 keypoint 1, the neck) is moved to the origin, and all other points are moved relative to it. To eliminate scale differences caused by varying heights and body proportions, a \textit{$1 \times 1$} normalization is then applied to these origin-shifted keypoints. In this scheme, the keypoints are scaled so that all points fit within a unit square of size $1\times1$. This is achieved by proportionally stretching the keypoints horizontally and vertically such that the maximum horizontal and vertical distances between any two points become exactly 1. The scaling factors are the reciprocals of these maximum distances. As shown in Figure~\ref{fig:1x1normapelda}, the transformed poses preserve the essential spatial relationships crucial for gesture recognition. This normalization method provides a generic solution to all the aforementioned challenges.

The final stage of our pipeline is the recognition model. This model takes the normalized keypoint data as input and performs the gesture classification. A detailed description of the recognition model is provided in the subsequent section.

\subsection{Sliding Window}
\label{ss:sliding_window}
Recognizing dynamic gestures requires analyzing a sequence of frames to determine the meaning of the gesture. A sliding window of 50 frames is used on the input sequence. According to \cite{tdk2024} this is the ideal length and it is naturally sufficient for recognizing dynamic gestures. With the 30 FPS camera used, this value corresponds to 1.67 seconds of input, an interval generally adequate for human recognition as well.

\subsection{Recognition Model}
\label{ss:dynamic_gesture_recognition}

%\subsection{Sliding Window}
%\label{ss:sliding_window}
%As established earlier, recognizing dynamic gestures requires analyzing a sequence of frames to determine the gesture's meaning.

%A sliding window of 50 frames is used on the input sequence.  According to \cite{tdk2024} 
%, a window size of approximately 50 frames yields the best results and 
%this is the ideal length and it is naturally sufficient for recognizing dynamic gestures. With the 30 FPS camera used, this value corresponds to 1.67 seconds of input, an interval generally adequate for human recognition as well.

%\subsection{Recognition Model Architecture}
%\label{ss:used_recognition_model}

%To evaluate the performance of different approaches, I use a fixed model architecture. This allows for a fair comparison, after which the best-performing method can be selected and the model optimized further.

The content of the sliding window is fed to a model consisting of 2 linear layers, followed by a GRU recurrent neural network \cite{chung2014empiricalevaluationgatedrecurrent}, headed by another 2 linear layers: 

\begin{enumerate}[nosep]
    \item \textbf{Linear} - N \\ $\rightarrow$ 2048
    \item \textbf{Linear} - 2048 \\ $\rightarrow$ 1024
    \item \textbf{GRU} - 1024 \\ $\rightarrow$ 256
    \item \textbf{Linear} - 256 \\ $\rightarrow$ 128
    \item \textbf{Linear} - 128 \\ $\rightarrow$ M (Number of output gestures)
\end{enumerate}

\subsection{Feature Representation Approaches}
\label{s:feature_representation_approaches}

Keypoints are extracted from each frame in the sliding window before feeding it into the recognition model. This section discusses alternatives considered in this study.

\subsubsection{Coordinate-based Approach}
\label{ss:coordinate_based_approach}

The raw data is a sequence of feature vectors $v_k = (x_{1,k}, y_{1,k}, \dots, x_{n,k}, y_{n,k})$ for each frame within the sliding window $k \in \{1, 2, \dots, 50\}$. Here, $(x_{i,k}, y_{i,k})$ represents the normalized coordinates of the $i$-th keypoint, where $i \in \{1, 2, \dots, n\}$. For arm gesture recognition, it is sufficient to consider  upper body keypoints only. Thus, keypoints 0..8 of the BODY-25 model are used ($n=9$).

%As previously discussed, applying one of the normalization methods (Section \ref{ss:normalization}) to these coordinates is essential to ensure invariance to scale and position.

\subsubsection{Angle-based Approach}
\label{ss:angle_based_approach}

\begin{figure}[htbp]
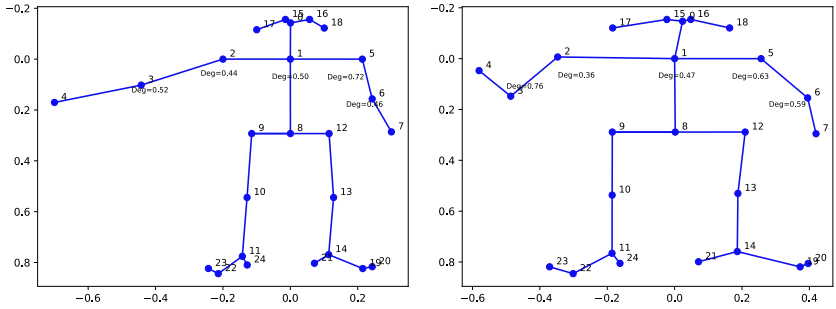

  \centering
  \includesvg[width=0.4\columnwidth]{images/szogek/20250511-222051-38795.svg}
  \includesvg[width=0.4\columnwidth]{images/szogek/20250511-222052-95674.svg}
  \caption{The various angles extracted (after normalization) and their values on two example frames.}
  \label{fig:dinamikusKarjelzesekSzoge}
\end{figure}

This method represents poses using the angles formed by relevant skeletal keypoints. The feature vector for each frame in the sliding window consists of these calculated angles.
The key angles considered are calculated based on the following keypoint sequences (where the middle number is the vertex):
\begin{enumerate}[nosep]
    \item Left elbow angle (2-3-4)
    \item Right elbow angle (5-6-7)
    \item Left shoulder angle (1-2-3)
    \item Right shoulder angle (1-5-6)
    \item Neck angle (0-1-8)
\end{enumerate}
The values of these angles are then transformed to the interval $[0,1]$.
This representation aims to capture the pose configuration independent of absolute positions or overall body orientation.
Examples are visualized in Figure~\ref{fig:dinamikusKarjelzesekSzoge}.

\section{Data and Methods}
\label{sec:data-and-methods}

\subsection{Dataset}
\label{sec:dataset}

A custom dataset was created, consisting of samples of 8, traffic-related dynamic arm signals (gestures).\footnote{The dataset also contains samples of the \texttt{StandStill} gesture, which represents a natural standing pose and is technically a static gesture.}
The gestures included are:
\begin{itemize}[nosep]
    \item \texttt{RightHandLeftCircle}: The right hand performs a counter-clockwise circular motion.
    \item \texttt{RightHandRightCircle}: The right hand performs a clockwise circular motion.
    \item \texttt{StandStill}: The subject maintains a neutral, natural standing posture.
    \item \texttt{LeftHandWave}: The left hand performs an up-and-down waving motion.
    \item \texttt{RightHandWave}: The right hand performs an up-and-down waving motion.
    \item \texttt{CallToPass}: The left arm is extended horizontally, and the right hand performs a beckoning/waving motion to signal "proceed" or "come through".
    \item \texttt{LeftHandRightCircle}: The left hand performs a clockwise circular motion.
    \item \texttt{LeftHandLeftCircle}: The left hand performs a counter-clockwise circular motion.
\end{itemize}
%There are 8 distinct dynamic gestures included.
For comprehensive testing of the model's robustness to viewpoint variations, additional real-world samples were recorded with subjects rotated left and right by 15, 30, and 45 degrees. These samples were exclusively used for the test set.
%\textcolor{red}{Description of the signals. At least names.}

\subsection{Training and Testing}
\label{ss:training_and_testing}

An extensive set of experiments was performed studying the various alternatives proposed for the pipeline components. During these experiments the dataset was split into training (60\%), validation (10\%), and testing (30\%) subsets.
Training was conducted using the Adam optimization algorithm with the CrossEntropy loss function.

It is important to emphasize that during initial training and validation, only data recorded with subjects facing the camera (frontal-view) was used. Real-world data involving subjects rotated left or right %(i.e., the samples mentioned in the dataset description above) 
(see the end of Section~\ref{sec:dataset}) were strictly reserved for the test set and excluded from training and validation. However, as detailed in Section \ref{s:rotated_training_data_generation}, artificially generated rotated data, derived from the frontal-view samples, was utilized in some training experiments to enhance model robustness.

\subsubsection{Experiments with Varying Speeds}
The impact of various gesture execution speeds on recognition accuracy was also studied. The training phase of these experiments was the same as before, using the normal recorded speed of the samples. 

During the test phase, the test dataset was modified to account for various speed levels. Linear interpolation between frames was applied to simulate gestures performed at different speeds. For the angle-based approach, this involved recalculating angles based on interpolated keypoint data.
The following speed ratios (to the original speed) were tested: 0.5, 0.75, 0.9, 1.0 (original), 1.1, 1.3, and 2.0. This range covers speeds from half to double the normal execution rate, encompassing most human-interpretable variations.

%\section{Rotated Training Data Generation}
\subsection{Data Augmentation}
\label{s:rotated_training_data_generation}

As observed in preliminary experiments (and discussed in \cite{tdk2024}), simple recognition models might struggle with gestures performed at various angles relative to the camera. Collecting real-world data for every possible orientation can be time-consuming and costly, especially for systems designed to recognize a large vocabulary of gestures. To address this challenge, an augmented dataset was also created consisting of artificially  rotated samples.

%\subsection{Motivation for Artificial Data}
\label{ss:motivation_artificial_data}
The primary motivation is to enhance the model's robustness to viewpoint variations without requiring extensive real-world data collection for every angle. Synthetically generated rotated samples (created from existing frontal-view data) may potentially improve generalization ability to unseen viewpoints. This is particularly beneficial when dealing with complex gestures or having limited data acquisition resources.

\subsubsection{Generation Process}
\label{ss:generation_process}
The process leverages the available frontal-view ($0^\circ$) data to create artificial samples simulating rotated views.

\begin{enumerate}
    \item \textbf{Depth Estimation:} Since  a standard 2D camera is used, depth information (Z-coordinate) is not directly available. The relative depth of relevant keypoints (specifically, BODY-25 model arm keypoints 2 through 7) is estimated based on typical human posture during the gesture. This estimation transforms the 2D keypoints into a pseudo-3D skeleton relative to a reference point (BODY-25 keypoint 1, the neck, as defined in Section \ref{ss:normalization}). These estimations are manually defined based on anatomical plausibility for each gesture type.
    \item \textbf{3D Rotation:} The resulting pseudo-3D point cloud representing the skeleton is then mathematically rotated around a vertical axis passing through the reference point (BODY-25 keypoint 1). We generate rotations for specific angles, such as $15^\circ$, $30^\circ$, and $45^\circ$, both left and right.
    \item \textbf{2D Projection:} After rotation, the Z-coordinates are discarded, projecting the rotated 3D points back onto a 2D plane. This yields the artificial 2D keypoint data representing the gesture as seen from the simulated angle.
\end{enumerate}

This method enables the generation of diverse viewpoint data from a limited set of frontal-view recordings using standard 2D cameras. Depth cameras could directly provide 3D data, but they are expensive, more complex to handle, and may have range limitations unsuitable for certain applications. The proposed augmentation approach avoids these drawbacks and works with readily available standard and cheap cameras.

Table \ref{tab:depth_estimates} shows example manual depth estimates for keypoints 2..7 for a few gestures. Values are relative to the neck (keypoint 1),  positive values indicating points farther from the camera (deeper) and negative values indicating points closer. While these estimates are not  accurate, they approximate the real 3D structure sufficiently to generate useful rotated views. These artificially generated rotated samples may augment the original training dataset.

\begin{table}[htbp]
    \centering
    \caption{Example manual depth estimates for keypoints relative to the neck.}
    \label{tab:depth_estimates}
    \begin{tabular}{|p{0.25\columnwidth}|c|c|c|c|c|c|}
        \hline
        \textbf{Gesture} & \textbf{KP2} & \textbf{KP3} & \textbf{KP4} & \textbf{KP5} & \textbf{KP6} & \textbf{KP7} \\
        \hline
        \small{\texttt{StandStill}} & 0 & 0.1 & -0.1 & 0 & 0.1 & -0.1 \\
        \hline
        \small{\texttt{LeftHandWave}} & 0 & -0.4 & -0.4 & 0 & 0.1 & -0.1 \\
        \hline
        \small{\texttt{LeftHandLeftCircle}} & 0 & -0.1 & -0.1 & 0 & 0.1 & -0.1 \\
        \hline
    \end{tabular}
\end{table}

\subsection{Real-time Recognition Considerations}
\label{s:real_time_recognition}

Implementing real-time gesture recognition introduces challenges not present in offline experiments. 
%This section briefly discusses key practical considerations.
%\subsection{Sliding Window Management}
%In real-time systems, 
Since data is received  frame by frame, the sliding window size must strike a good balance between accuracy and speed. For smooth recognition, it's better to keep some old frames (e.g., half the window) and add new ones, rather than starting fresh each time or changing only one frame. This helps keep the recognition consistent and fast.

%\subsection{Adapting to FPS Variations}
Also, different cameras have different frame rates (FPS). In the experiments reported here a 50-frame window was used, expecting about 30 FPS. However, the window size should change with varying FPS rate. This makes sure the system looks at the same amount of time for a gesture as it did in training.

%\subsection{Output Smoothing}
In order to avoid the "flickering" of the recognised gesture (i.e., switching between recognized gestures too quickly) several recent outputs may be used.  A simple approach is to pick the gesture that appears most often in the last $N$ results. This makes the output more stable and ignores occasional errors.

%\section{Results of Dynamic Gesture Recognition}
\section{Results}
\label{s:results}

\begin{figure}[htbp]
  \centering
  \includesvg[width=\columnwidth]{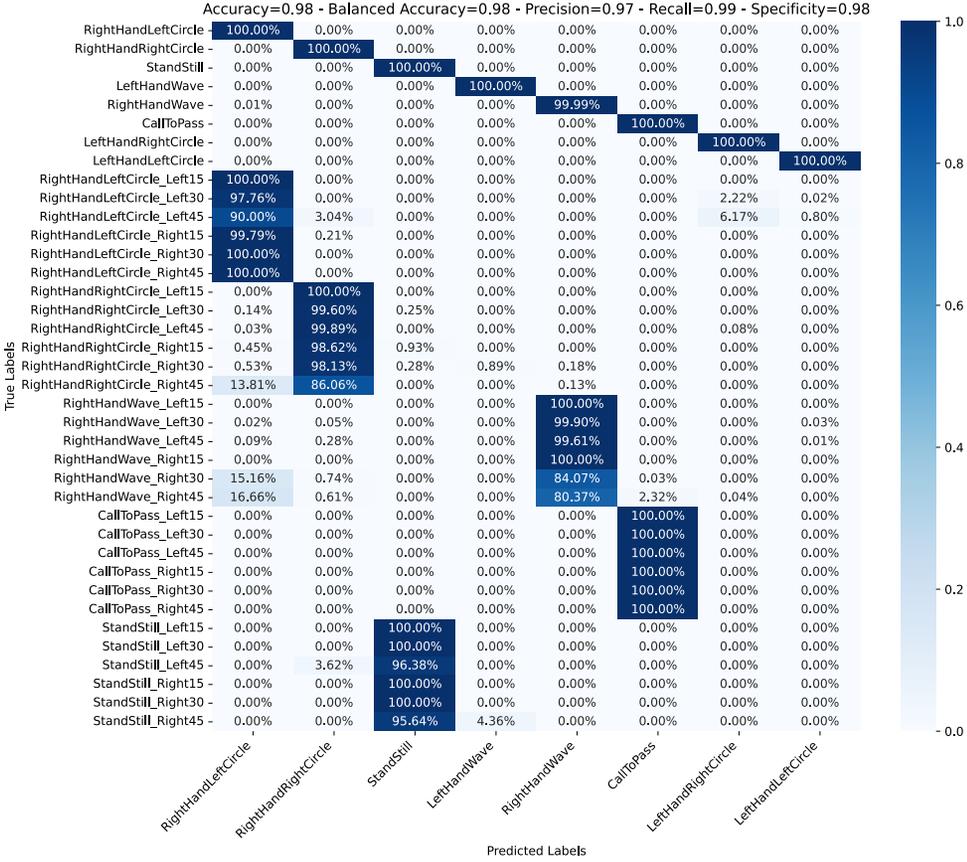}
  \caption{Successful recognition of rotated samples (coordinate-based approach with \textit{$1 \times 1$} normalization)}
  \label{fig:dinamikusKoord1x1}
\end{figure}

\begin{figure}[htbp]
  \centering
  \includesvg[width=\columnwidth]{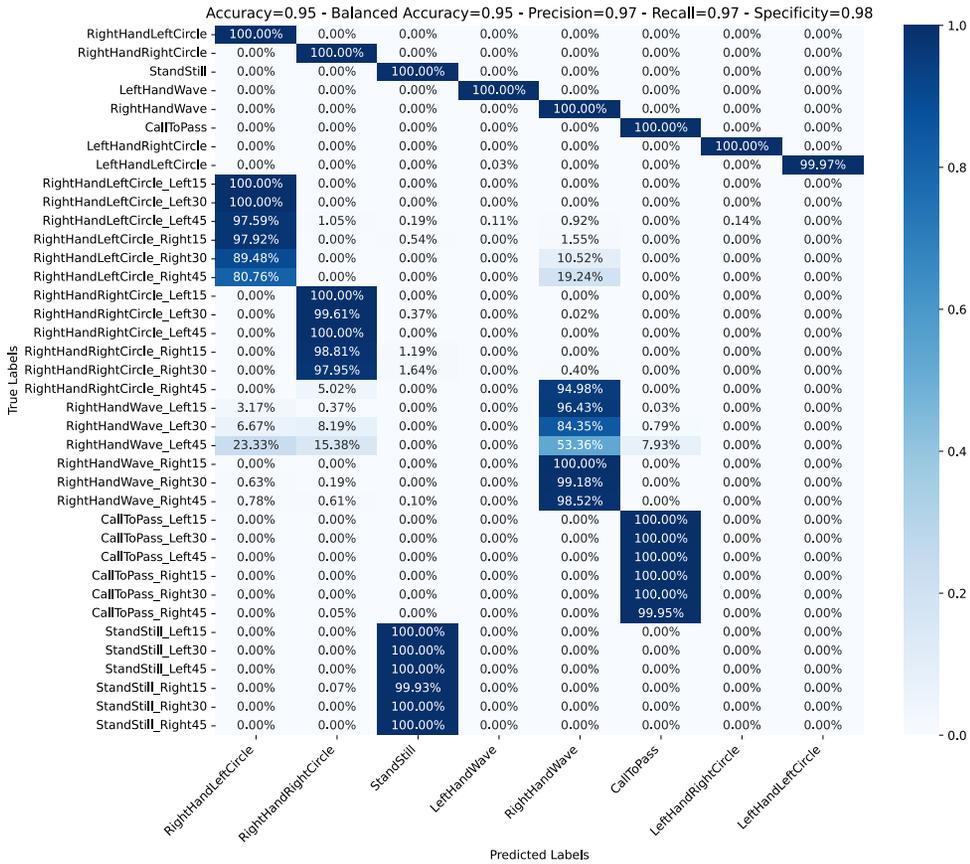}
  \caption{Successful recognition of rotated samples 
 (angle-based approach without normalization)}
  \label{fig:dinamikusSzogNormalizalasNelkul}
\end{figure}

This section presents the  results of our experiments. 

\subsection{Gestures Viewed from an Angle}

Performance on test sets containing gestures recorded from an angle is evaluated first. Both 
the coordinate-based and angle-based approaches perform well when trained on the augmented dataset containing  artificially generated rotated data. Figures~\ref{fig:dinamikusKoord1x1}~and~\ref{fig:dinamikusSzogNormalizalasNelkul} show the confusion matrices of the recognition results of the coordinate-based approach (with $1 \times 1$ normalization) and the angle-based approach (without normalization), respectively. Each row corresponds to the true label of the gesture (e.g., \texttt{LeftHandWave}, viewed from $-30^\circ$), while each column represents the recognition output. Dark blue cells denote a near 100\% rate, while white marks  0\%. There are two remarkably different parts of both matrices. The top quarter contains non-rotated inputs. Here dark blue cells in the diagonal are expected. The bottom part of the matrices contains samples viewed from an angle, grouped by  gesture. Here good performance is marked dark blue columns as gestures get classified in the right column independent of the view angle.  
While both methods are robust, the coordinate-based approach tends to yield better results at larger rotation angles, whereas the angle-based approach can offer faster and more accurate recognition at smaller rotation angles.

\subsection{Sensitivity to Gesture Speed}
\label{sec:sensitivity-to-speed}

%\subsubsection{Effectiveness with Different Speeds}

\begin{figure}[htbp]
  \centering
  \includesvg[width=\columnwidth]{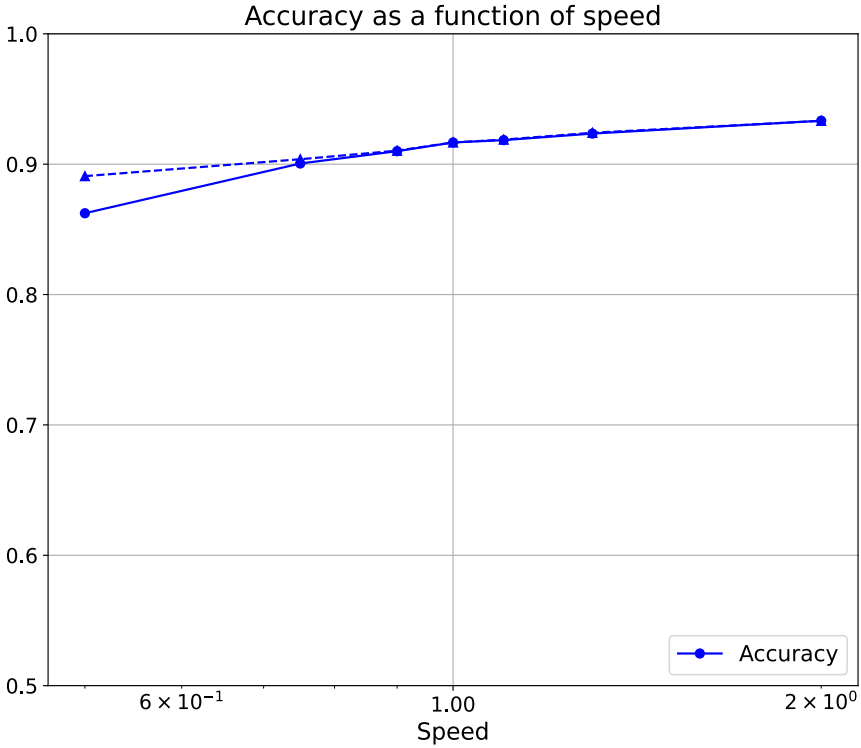}
  \caption{Accuracy as a function of speed, with speed-adjusted sliding window size. The results shown are for the angle-based feature representation (see Section~\ref{ss:angle_based_approach}) without applying additional normalization to the angle values.}
  \label{fig:sebessegplot_korrigalt}
\end{figure}

Figure \ref{fig:sebessegplot_korrigalt} illustrates the performance of the pipeline at various speed ratios. These experiments were conducted using a single, fixed recognition model to ensure that only the gesture execution speed varied as a factor. The dashed lines show experiments when the size of the sliding window was  adjusted proportionally to the gesture speed (e.g., a 100-frame window for 0.5x speed, or a 25-frame window for 2.0x speed). The results indicate that adapting the window size helps stabilizing accuracy across varying speeds.

Overall, the impact of gesture speed on recognition is  minimal. The proposed pipeline exhibits a high level of robustness against  variations in execution speed. There is a slight trend of faster gestures yielding marginally better accuracy (and slower ones underperforming), these differences are negligible for practical use. Even for  gestures  performed at half or double of the original speed, the accuracy deviation is minor.

%This confirms that the model (potentially referring to one detailed in Section \ref{ss:dinamikusBEST}, if defined) 

\section{Speed Measurement}

The results of Section~\ref{sec:sensitivity-to-speed} show that the proposed pipeline is largely insensitive to changes in the speed of the arm signal. However, as discussed in Section~\ref{s:introduction}, speed may also be a factor in determining the meaning of signals. 
%\textit{For example,  here is fast/slow circling or fast/slow waving. In both cases, the speed of the signal can have important meaning.}
If this is the case, signal speed must also be calculated. This section discusses this problem and proposes a solution.

\subsection{Problem Statement}
%As we saw the recognition accuracy of the selected approach is essentially independent of the speed of the issued signal.
%Itt majd hivatkozzuk meg, hogy ez igazából a szöges megközleítés

%This sounds good for signal recognition, as in real applications, signals are not always issued at the same speed as during training.
%However, recognize that the meaning of dynamic signals can also be influenced by the speed at which the signal is issued.
%If we want to distinguish between dynamic signals based on speed, 

A general approach to distinguish between signals based on their speeds is to assign a numerical value to the current content of the sliding window that characterizes the speed of the signal. Various speed categories (fast,  slow, etc.) can be easily and flexibly derived from this value.

\subsection{The Proposed Approach}

The approach proposed takes advantage of the fact that dynamic signals are predominantly cyclic. That is, the signal consists of certain movements performed repeatedly.

\begin{figure}[htbp]
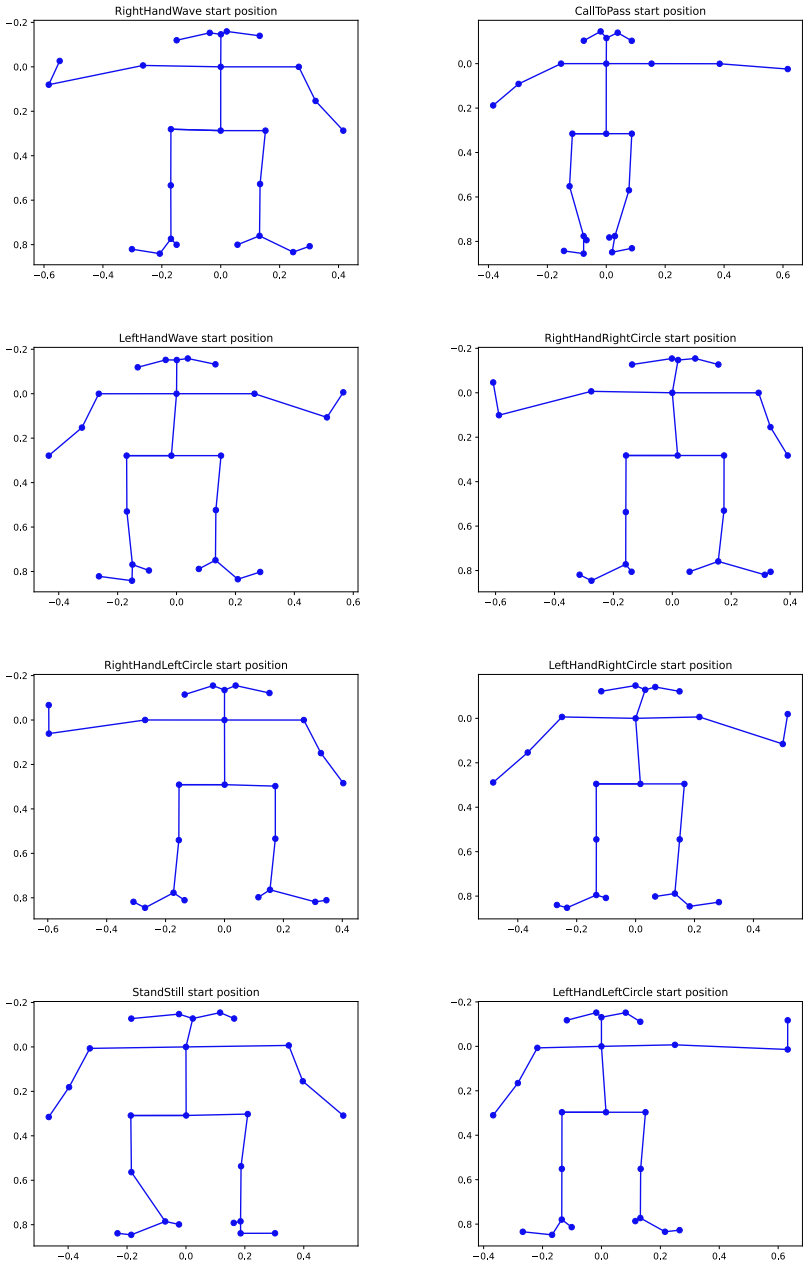

  \centering
    \begin{tabular}{cc}
    \includesvg[width=0.4\columnwidth]{images/sebessegmeres/20250513-225010-20052.svg} &
    \includesvg[width=0.4\columnwidth]{images/sebessegmeres/20250513-225010-45071.svg} \\
    \includesvg[width=0.4\columnwidth]{images/sebessegmeres/20250513-225010-49457.svg} &
    \includesvg[width=0.4\columnwidth]{images/sebessegmeres/20250513-225010-50517.svg} \\
    \includesvg[width=0.4\columnwidth]{images/sebessegmeres/20250513-225010-67302.svg} &
    \includesvg[width=0.4\columnwidth]{images/sebessegmeres/20250513-225010-82696.svg} \\
    \includesvg[width=0.4\columnwidth]{images/sebessegmeres/20250513-225010-94149.svg} &
    \includesvg[width=0.4\columnwidth]{images/sebessegmeres/20250513-225010-96739.svg} \\
    \end{tabular}
    \caption{Selected start positions for various gestures.}
    \label{fig:kivalasztottnullapoziciok}
\end{figure}

Taking advantage of this observation, a \textit{start position} (pose) is selected for each signal. This pose is  characteristic of the given dynamic signal, such that giving  the signal, this pose is  periodically revisited (i.e., a pose sufficiently similar to the start position appears). The \textit{start positions} of the traffic signal dataset are shown on Figure~\ref{fig:kivalasztottnullapoziciok}.

\begin{figure}[htbp]
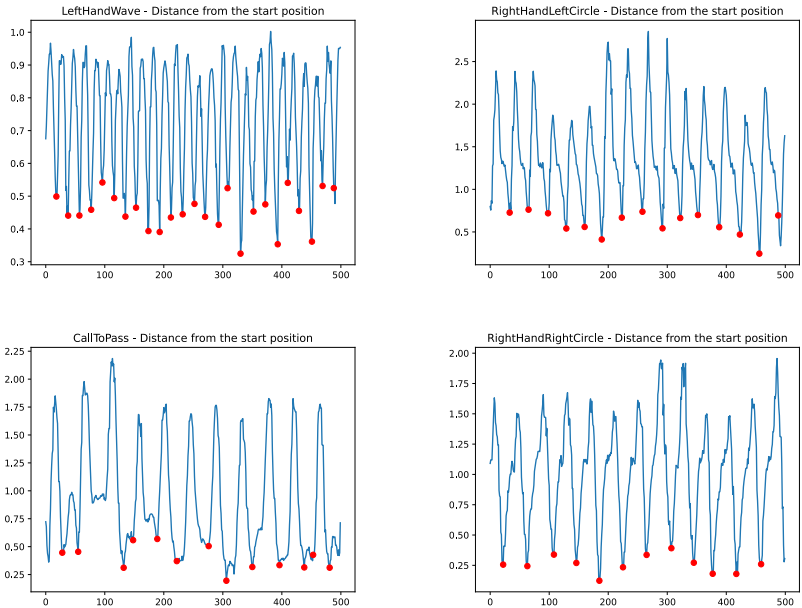

  \centering
  \begin{tabular}{cc}
  \includesvg[width=0.4\columnwidth]{images/sebessegmeres/balMeneszttavolsagok_P.svg} &
  \includesvg[width=0.4\columnwidth]{images/sebessegmeres/balraKoroztavolsagok_P.svg} \\
  \includesvg[width=0.4\columnwidth]{images/sebessegmeres/felhivasAthaladasratavolsagok_P.svg} &
  \includesvg[width=0.4\columnwidth]{images/sebessegmeres/jobbraKoroztavolsagok_P.svg} \\
  \end{tabular}
  \caption{Distances from selected start positions for different gestures, with identified local minima marked by red dots.}
  \label{fig:eltavolsagok}
\end{figure}

The  results of Section~\ref{sec:sensitivity-to-speed} indicate that dynamic signals can be recognized independently of speed.  Therefore,  after signal  recognition is performed,  the deviation (practically: distance) from the selected \textit{start position} is calculated for each frame (pose) in the  sliding window (see Figure~\ref{fig:eltavolsagok}).
From here, a measure of speed can be determined. 

Speed is understood here as the  distance (number of frames) between the first two  occurrences of the \textit{start position} in the examined sliding window.\footnote{This approach discretizes the problem and, given the FPS rate, can be directly linked to wall-clock time.} Given the time series (per frame) of distances from the \textit{start position} discussed above, this problem is equivalent to finding the first two \textit{local minima} of the series (marked by red dots in the right column of Figure~\ref{fig:eltavolsagok}).

%\subsection{Speed Concept}

%A legitimate question might arise as to what we mean by speed and what this speed should measure.
%We might expect some real-world elapsed time as speed.
%\textit{How many milliseconds did it take to complete one period of the signal?}

%This is also good because it reduces the events occurring in continuous time to a well-discretized problem, and a discrete task is typically easier to work with.
%\textit{How many states (frames) elapsed between the two \textit{start positions}?}

%Ezek a nulla pozíciók nem biztos, hogy annyira fontosak, hogy foglalják a helyet, de mutatósak

%ezeket a nulllától eltelt távolságokat mindenképp beletenném, mert innen a piros pontok miatt könnyen látszik a sebességfogalom is...

%\subsection{Speed Measurement Analytically}
%\subsection{Speed Calculation}

%Speed can also be calculated analytically, based on the deviations between \textit{start positions}.

%Of course, the local minima shown in Figure~\ref{fig:eltavolsagok} must be found first, marked with red dots. Naturally, the speed is the distance between two consecutive red dots.

Local minima can be determined using various methods.
One can search for local minima by examining a few neighboring points (recall that the problem is discrete). For instance, in Figure~\ref{fig:eltavolsagok}, the red dots marking local minima were identified by iterating through the sequence of distance values and selecting points that are smaller than some of their neighbors.
Alternatively, it is also possible to solve an interpolation task defined by the points and then find the local minima by deriving the obtained interpolation polynomial.

\section{Conclusions}

This paper presented a real-time pipeline for dynamic arm gesture recognition based on OpenPose keypoint estimation, keypoint normalization, and a recurrent neural network classifier. A specific normalization  scheme, two feature representations (coordinate- and angle-based) and a data augmentation approach were also discussed. Experiments on a custom traffic-control gesture dataset demonstrated high recognition accuracy across varying viewing angles and speeds. % Finally, the approach is validated by an application where a TurtleBot 3 is controlled by arm signals.
Therefore the proposed pipeline is  robust not only against  changes in viewing angle but also against variations in gesture speed. This is crucial for real-world applications where viewing angles and gesture speeds may naturally vary.
Finally, a solution to calculate the speed of the dynamic arm signal (if meaningful) was also outlined.

\begin{acks}
Supported by the EKÖP-24 University Excellence Scholarship Program of the Ministry for Culture
and Innovation from the source of the National Research, Development and Innovation Fund.

%The authors would also like to acknowledge the support of ELTE Eötvös Loránd University, Budapest, Hungary.
\end{acks}

\bibliographystyle{ACM-Reference-Format}
\bibliography{introb25}


\begin{thebibliography}{8}

%%% the Web address.

\ifx \showCODEN    \undefined \def \showCODEN     #1{\unskip}     \fi
\ifx \showISBNx    \undefined \def \showISBNx     #1{\unskip}     \fi
\ifx \showISBNxiii \undefined \def \showISBNxiii  #1{\unskip}     \fi
\ifx \showISSN     \undefined \def \showISSN      #1{\unskip}     \fi
\ifx \showLCCN     \undefined \def \showLCCN      #1{\unskip}     \fi
\ifx \shownote     \undefined \def \shownote      #1{#1}          \fi
\ifx \showarticletitle \undefined \def \showarticletitle #1{#1}   \fi
\ifx \showURL      \undefined \def \showURL       {\relax}        \fi
% invisible to TeX
\providecommand\bibfield[2]{#2}
\providecommand\bibinfo[2]{#2}
\providecommand\natexlab[1]{#1}
\providecommand\showeprint[2][]{arXiv:#2}

\bibitem[Cao et~al\mbox{.}(2019)]%
        {cao2019openposerealtimemultiperson2d}
\bibfield{author}{\bibinfo{person}{Zhe Cao}, \bibinfo{person}{Gines Hidalgo},
  \bibinfo{person}{Tomas Simon}, \bibinfo{person}{Shih-En Wei}, {and}
  \bibinfo{person}{Yaser Sheikh}.} \bibinfo{year}{2019}\natexlab{}.
\newblock \showarticletitle{OpenPose: Realtime Multi-Person 2D Pose Estimation
  using Part Affinity Fields}.
\newblock  (\bibinfo{year}{2019}).
\newblock
\showeprint[arxiv]{1812.08008}~[cs.CV]
\urldef\tempurl%
\url{https://arxiv.org/abs/1812.08008}
\showURL{%
\tempurl}


\bibitem[Chung et~al\mbox{.}(2014)]%
        {chung2014empiricalevaluationgatedrecurrent}
\bibfield{author}{\bibinfo{person}{Junyoung Chung}, \bibinfo{person}{Caglar
  Gulcehre}, \bibinfo{person}{KyungHyun Cho}, {and} \bibinfo{person}{Yoshua
  Bengio}.} \bibinfo{year}{2014}\natexlab{}.
\newblock \showarticletitle{Empirical Evaluation of Gated Recurrent Neural
  Networks on Sequence Modeling}.
\newblock  (\bibinfo{year}{2014}).
\newblock
\showeprint[arxiv]{1412.3555}~[cs.NE]
\urldef\tempurl%
\url{https://arxiv.org/abs/1412.3555}
\showURL{%
\tempurl}


\bibitem[He et~al\mbox{.}(2020)]%
        {HE2020248}
\bibfield{author}{\bibinfo{person}{Jian He}, \bibinfo{person}{Cheng Zhang},
  \bibinfo{person}{Xinlin He}, {and} \bibinfo{person}{Ruihai Dong}.}
  \bibinfo{year}{2020}\natexlab{}.
\newblock \showarticletitle{Visual Recognition of traffic police gestures with
  convolutional pose machine and handcrafted features}.
\newblock \bibinfo{journal}{\emph{Neurocomputing}}  \bibinfo{volume}{390}
  (\bibinfo{year}{2020}), \bibinfo{pages}{248--259}.
\newblock
\showISSN{0925-2312}
\href{https://doi.org/10.1016/j.neucom.2019.07.103}{doi:\nolinkurl{10.1016/j.neucom.2019.07.103}}


\bibitem[Liu et~al\mbox{.}(2021)]%
        {app112411951}
\bibfield{author}{\bibinfo{person}{Kang Liu}, \bibinfo{person}{Ying Zheng},
  \bibinfo{person}{Junyi Yang}, \bibinfo{person}{Hong Bao}, {and}
  \bibinfo{person}{Haoming Zeng}.} \bibinfo{year}{2021}\natexlab{}.
\newblock \showarticletitle{Chinese Traffic Police Gesture Recognition Based on
  Graph Convolutional Network in Natural Scene}.
\newblock \bibinfo{journal}{\emph{Applied Sciences}} \bibinfo{volume}{11},
  \bibinfo{number}{24} (\bibinfo{year}{2021}).
\newblock
\showISSN{2076-3417}
\href{https://doi.org/10.3390/app112411951}{doi:\nolinkurl{10.3390/app112411951}}


\bibitem[Ma et~al\mbox{.}(2018)]%
        {ijgi7010037}
\bibfield{author}{\bibinfo{person}{Chunyong Ma}, \bibinfo{person}{Yu Zhang},
  \bibinfo{person}{Anni Wang}, \bibinfo{person}{Yuan Wang}, {and}
  \bibinfo{person}{Ge Chen}.} \bibinfo{year}{2018}\natexlab{}.
\newblock \showarticletitle{Traffic Command Gesture Recognition for Virtual
  Urban Scenes Based on a Spatiotemporal Convolution Neural Network}.
\newblock \bibinfo{journal}{\emph{ISPRS International Journal of
  Geo-Information}} \bibinfo{volume}{7}, \bibinfo{number}{1}
  (\bibinfo{year}{2018}).
\newblock
\showISSN{2220-9964}
\href{https://doi.org/10.3390/ijgi7010037}{doi:\nolinkurl{10.3390/ijgi7010037}}


\bibitem[Mishra et~al\mbox{.}(2021)]%
        {s21237914}
\bibfield{author}{\bibinfo{person}{Ashutosh Mishra}, \bibinfo{person}{Jinhyuk
  Kim}, \bibinfo{person}{Jaekwang Cha}, \bibinfo{person}{Dohyun Kim}, {and}
  \bibinfo{person}{Shiho Kim}.} \bibinfo{year}{2021}\natexlab{}.
\newblock \showarticletitle{Authorized Traffic Controller Hand Gesture
  Recognition for Situation-Aware Autonomous Driving}.
\newblock \bibinfo{journal}{\emph{Sensors}} \bibinfo{volume}{21},
  \bibinfo{number}{23} (\bibinfo{year}{2021}).
\newblock
\showISSN{1424-8220}
\href{https://doi.org/10.3390/s21237914}{doi:\nolinkurl{10.3390/s21237914}}


\bibitem[Sathya and Geetha(2015)]%
        {SATHYA20151700}
\bibfield{author}{\bibinfo{person}{R. Sathya} {and}
  \bibinfo{person}{M.~Kalaiselvi Geetha}.} \bibinfo{year}{2015}\natexlab{}.
\newblock \showarticletitle{Framework for Traffic Personnel Gesture
  Recognition}.
\newblock \bibinfo{journal}{\emph{Procedia Computer Science}}
  \bibinfo{volume}{46} (\bibinfo{year}{2015}), \bibinfo{pages}{1700--1707}.
\newblock
\showISSN{1877-0509}
\href{https://doi.org/10.1016/j.procs.2015.02.113}{doi:\nolinkurl{10.1016/j.procs.2015.02.113}}
\newblock
\shownote{Proceedings of the International Conference on Information and
  Communication Technologies, ICICT 2014, 3-5 December 2014 at Bolgatty Palace
  Island Resort, Kochi, India}.


\bibitem[Zsolt(2024)]%
        {tdk2024}
\bibfield{author}{\bibinfo{person}{Bagladi~Milán Zsolt}.}
  \bibinfo{year}{2024}\natexlab{}.
\newblock \showarticletitle{Mesterséges Intelligencia felhasználása emberi
  karjelzések értelmezésében}.
\newblock  (\bibinfo{year}{2024}).
\newblock


\end{thebibliography}

\end{document}